\documentclass{article}
\usepackage{tracesampler_preprint,times}

\usepackage{microtype}
\usepackage{amsmath,amssymb,amsfonts}
\usepackage{graphicx}
\usepackage{booktabs}
\usepackage{multirow}
\usepackage{array}
\usepackage{placeins}
\usepackage{float}
\usepackage{hyperref}
\hypersetup{
    colorlinks=true,
    linkcolor=blue,
    citecolor=blue,
    urlcolor=blue,
    pdfauthor={George Matthew, Guoqiang Zhang, Kenta Niwa}
}
\usepackage{url}
\usepackage{comment}

\makeatletter
\def\section{\@startsection{section}{1}{\z@}%
  {-1.20ex}%
  {0.60ex}%
  {\large\sc\raggedright}}
\def\subsection{\@startsection{subsection}{2}{\z@}%
  {-0.95ex}%
  {0.30ex}%
  {\normalsize\sc\raggedright}}

\def\paragraph{\@startsection{paragraph}{4}{\z@}%
  {0.50ex}%
  {-1em}%
  {\normalfont\normalsize\bfseries}}
\makeatother

\AtBeginDocument{%
  \setlength{\abovedisplayskip}{3pt}%
  \setlength{\belowdisplayskip}{3pt}%
  \setlength{\abovedisplayshortskip}{3pt}%
  \setlength{\belowdisplayshortskip}{3pt}%
}

\title{TraceSampler: Making Sampling Errors Predictable and Correctable for Accelerated Generative Sampling}

\author{%
  {\normalfont\bfseries George Matthew$^{1}$, Guoqiang Zhang$^{1}$, Kenta Niwa$^{2}$}\\[2pt]
  {\normalfont\small $^{1}$University of Exeter, United Kingdom \quad $^{2}$NTT Communication Science Laboratories, Japan}\\[1pt]
  {\normalfont\scriptsize\texttt{gm649@exeter.ac.uk}\quad\texttt{g.z.zhang@exeter.ac.uk}\quad\texttt{kenta.niwa@ntt.com}}%
}

\begin{document}

\maketitle

\begin{abstract}

Accurate sampling from continuous-time generative models can require many model evaluations. Low-step sampling reduces this cost, but coarse updates introduce endpoint errors that can accumulate along the sampling trajectory and degrade generation quality. We introduce TraceSampler, a post-training method that corrects these errors without modifying the pretrained model. Rather than directly optimising the low-step update itself, TraceSampler models its endpoint error as a sample-specific prediction problem. A higher-step reference integration over the same interval provides the supervised target, while a compact representation makes the high-dimensional error predictable from the current sampling trace. This representation combines shared residual structure, directions derived from the trace, and a low-rank remainder, so shallow per-step MLP heads need only predict sample-dependent coefficients. These coefficients drive a two-stage learned residual correction: the first stage predicts the full endpoint residual, while a backbone evaluation at the intermediate state supplies new trace information for predicting the residual that remains; this evaluation is reused by the following step. TraceSampler is evaluated with Rectified Flow on CIFAR-10, SiT-XL/2 on ImageNet-256, and SD3.5 Medium for text-to-image generation. At 10 steps, it reduces trajectory MSE relative to matched-step Euler by 90.5\% on CIFAR-10 and 85.6\% on ImageNet-256, while achieving FID scores of 2.692 and 2.189, close to 120-step Heun references of 2.601 and 2.056 with 24$\times$ fewer backbone evaluations. On SD3.5 Medium, TraceSampler improves image quality and prompt following; at 10 steps, ImageReward rises from 0.752 to 0.906 while trajectory MSE falls by 70.7\%. These gains preserve Euler's backbone-evaluation budget, add less than 0.4\% inference FLOPs, and incur negligible wall-clock overhead, showing that coarse-step sampling error can be made predictable from the sampling trace and efficiently corrected.

\end{abstract}

\begin{center}
    \begin{minipage}{0.82\linewidth}
        \begin{minipage}[t]{0.48\linewidth}
            \centering
            \includegraphics[width=\linewidth]{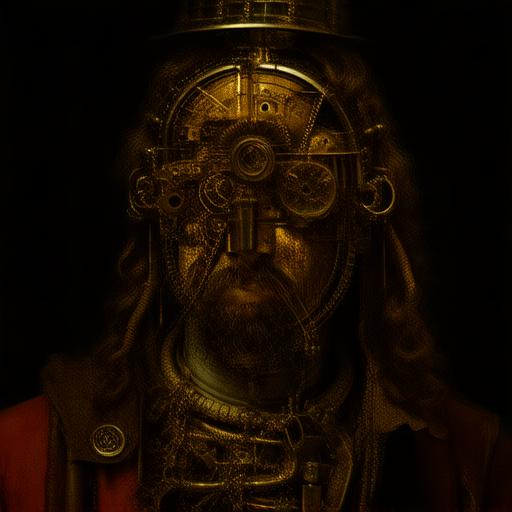}

            \smallskip
            \textbf{Euler, 8 steps}
        \end{minipage}
        \hfill
        \begin{minipage}[t]{0.48\linewidth}
            \centering
            \includegraphics[width=\linewidth]{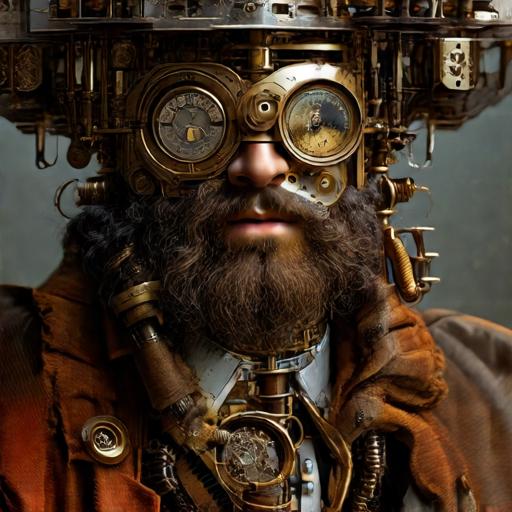}

            \smallskip
            \textbf{TraceSampler, 8 steps}
        \end{minipage}

        \medskip
        \footnotesize
        \textbf{SD3.5 Medium, \(512\times512\).}
        Prompt: \textit{``Portrait of Leonardo da Vinci as a steampunk cyborg,
        clockwork automaton with medieval technology.''}
    \end{minipage}
\end{center}

\section{Introduction}

Continuous-time generative models sample by numerically integrating a learned
ODE from noise to data \citep{lipman2022flow,song2020score}. Fewer steps
accelerate generation but require coarser updates whose endpoint errors
accumulate along the sampling trajectory and degrade generation quality. The
challenge is to approach high-step accuracy with far fewer model evaluations.

\begin{comment}
Existing accelerated samplers reduce few-step error mainly along three approaches: designing more accurate numerical update rules \citep{lu2025dpm,zhao2023unipc}, optimising solver parameters or sampling schedules, and fitting or learning a correction to the sampling trajectory on top of a base solver \citep{zhang2023accelerating,wang2025pas,wang2025differentiable}. DC-Solver \citep{zhao2024dcsolver} combines the first two approaches, augmenting a predictor--corrector solver with an optimised, per-step compensation ratio. Despite this variety, the update rule in each case is fixed in advance, and any correction parameters are fitted offline and then applied identically to every sample: none of these approaches adapts to the specific trajectory being sampled at inference.
\end{comment}

Existing accelerated samplers reduce few-step error through more accurate numerical updates, optimisation of solver parameters or sampling schedules, and fitted or learned corrections to the sampling trajectory \citep{lu2025dpm,zhao2023unipc,zhang2023accelerating,zhao2024dcsolver,wang2025pas,wang2025differentiable}. These methods typically improve the update rule itself or fit solver or correction parameters that are determined before inference and shared across samples. While effective, this does not explicitly account for how the error made by a low-step update may differ from one sampling trajectory to another. If these differences are systematic, using the same solver design or fitted correction across samples may leave trajectory-specific error uncorrected.

Our hypothesis is that the endpoint error of a low-step update contains systematic structure tied to the current sampling trace, and that a compact representation can make this structure predictable. We introduce \textbf{TraceSampler}, a post-training framework built around this hypothesis. During offline fitting, the deployed low-step update and a higher-step
reference integration start from the same state and span the same interval.
Their endpoint difference---the \emph{coarse-to-reference endpoint residual}---
defines the step-local error made by the deployed update and, equivalently, the
correction required to map its endpoint to the reference endpoint. TraceSampler
treats this error as a per-trajectory supervised prediction target; the
reference is absent at inference.

Because this error is image- or latent-dimensional, predicting it directly is
impractical. TraceSampler therefore represents it compactly using shared
residual structure, explicit directions derived from the current sampling
trace, and a low-rank component for remaining variation. Shallow per-step MLP
heads predict only the corresponding sample-dependent coefficients.

These coefficients reconstruct trajectory-specific residual estimates. The
residual predictor estimates the full endpoint error from information already
available in the sampling trace. A frozen-backbone evaluation at the
intermediate state then adds new information to the trace, which the residual
corrector uses to estimate the remaining error. This evaluation is reused by
the following Euler update, preserving the matched backbone-evaluation budget.

Across CIFAR-10, ImageNet-256, and SD3.5 Medium, TraceSampler substantially
improves low-step sampling. At 10 steps, it reaches FID 2.692 and 2.189 on
CIFAR-10 and ImageNet-256, close to 120-step Heun values of 2.601 and 2.056,
while reducing trajectory MSE by 90.5\% and 85.6\% with \(24\times\) fewer
backbone evaluations. On SD3.5 Medium, ImageReward rises from 0.752 to 0.906
while trajectory MSE falls by 70.7\%, with improvements in image quality,
prompt following, and estimated human preference relative to matched Euler. These gains preserve Euler's
backbone-evaluation budget, add below 0.4\% inference FLOPs, and incur negligible
measured wall-clock overhead.

\section{Related Work and Positioning}
\label{sec:positioning}

Few-step samplers reduce coarse-step error through more accurate numerical
updates, offline calibration of solver parameters or sampling schedules, and
fitted corrections to the sampling trajectory. Representative approaches include
DPM-Solver++ \citep{lu2025dpm} and UniPC \citep{zhao2023unipc}, which derive
numerical update rules; IIA \citep{zhang2023accelerating}, which fits step-wise
solver coefficients against finer integrations; and methods such as DC-Solver,
DSS, and PAS that optimise compensation, solver/timestep parameters, or
low-dimensional corrections \citep{zhao2024dcsolver,wang2025pas,wang2025differentiable}.

TraceSampler differs at the level of what is modelled. Rather than learning or
calibrating the sampling rule itself, it retains Euler as the deployed numerical
rule and models the realised same-interval endpoint error made by that rule.
This error is represented using shared residual structure, stage-specific
directions constructed from the current sampling trace, and a low-rank
remainder, with the corresponding coefficients predicted separately for each
trajectory. TraceSampler therefore performs structured, sample-dependent
prediction of the error made by the current update, rather than directly
modifying the integration rule or making the deployed correction depend solely
on quantities fixed offline and shared across trajectories.

Among the closest comparisons, UniPC also has a predictor--corrector structure,
but UniP and UniC are numerical state-update formulas, whereas TraceSampler's
residual predictor and residual corrector are learned estimators of error around
a fixed Euler update \citep{zhao2023unipc}. IIA is closely related through its
use of finer integrations offline, but uses them to fit solver coefficients that
are subsequently reused across samples, rather than to supervise per-trajectory
endpoint-error prediction \citep{zhang2023accelerating}. We therefore focus our
empirical comparison on IIA--Euler as a same-base fitted baseline and UniPC as
a strong matched-budget few-step solver, isolating the two closest comparisons.

The contributions are threefold: (i) casting the realised same-interval endpoint
error of each deployed low-step update as a supervised per-trajectory prediction
problem; (ii) a structured residual representation combining shared structure,
stage-specific trajectory directions, and a low-rank remainder with
sample-dependent coefficient prediction; and (iii) a two-stage learned residual
correction that uses an intermediate backbone evaluation to refine the remaining
error while reusing that evaluation in the following step, thereby preserving
the backbone-evaluation budget.

\section{TraceSampler}
\label{sec:tracesampler}

\subsection{Predicting and Correcting Coarse-Step Sampling Error}
\label{sec:prop1}

TraceSampler retains an Euler-form coarse update as its low-step backbone and learns the endpoint error made by that update. At step \(k\), let \(x_k\) be the corrected state at \(t_k\) and \(v_k\) the backbone velocity available to the sampler. The coarse endpoint is \(x_{k+1}^{\mathrm{base}}=x_k+\Delta t_kv_k\), where \(\Delta t_k=t_{k+1}-t_k\). With few sampling steps, this endpoint can deviate substantially from a higher-step reference over the same interval. During offline fitting, the reference solver starts from the same \(x_k\) and integrates over \([t_k,t_{k+1}]\), giving \(x_{k+1}^{\mathrm{ref}}\). Their endpoint difference defines the supervised residual target:
\[
r_k=x_{k+1}^{\mathrm{ref}}-x_{k+1}^{\mathrm{base}}.
\]

Thus, \(r_k\) is exactly the correction required to map the deployed coarse endpoint to its reference endpoint. The reference integration is used only to construct offline targets; it is absent at inference.

\begin{figure}[h]%[!htbp]
    \centering
    \includegraphics[width=0.9\linewidth]{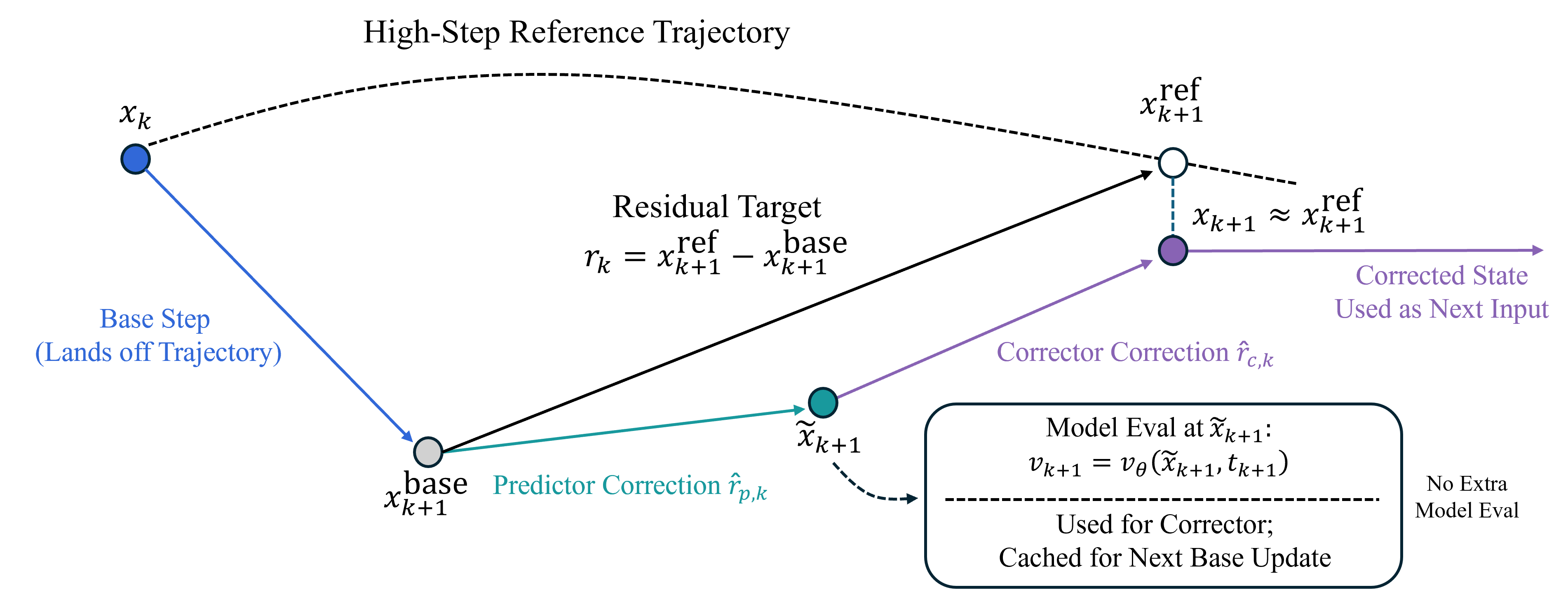}
    \caption{Two-stage learned residual correction for one sampling step.}
    \label{fig:two_stage_residual_geometry}
\end{figure}

TraceSampler leaves the base update unchanged and learns two residual-estimation stages around it. The residual predictor uses information already available at the start of the step to estimate the full coarse-step error and moves the base endpoint to an intermediate state \(\tilde x_{k+1}\). 
The frozen backbone is then evaluated at this intermediate state, \(v_{k+1}=v_\theta(\tilde x_{k+1},t_{k+1})\), adding new information to the sampling trace that the residual corrector uses to predict only the error left after the first correction. 
This evaluation is reused by the following step. Figure~\ref{fig:two_stage_residual_geometry} illustrates this sequence.

Concretely, the residual predictor targets the full coarse-to-reference residual and produces
\[
r_{p,k} = x_{k+1}^{\mathrm{ref}}-x_{k+1}^{\mathrm{base}}=r_k,
\qquad
\tilde x_{k+1} = x_{k+1}^{\mathrm{base}}+\hat r_{p,k},
\]
where \(\hat r_{p,k}\) is the predicted correction. Using the new backbone evaluation at \(\tilde x_{k+1}\), the residual corrector targets only the remaining gap,
\[
r_{c,k} = x_{k+1}^{\mathrm{ref}}-\tilde x_{k+1},
\qquad
x_{k+1} = \tilde x_{k+1}+\hat r_{c,k}.
\]

The intermediate-state evaluation is reused as the velocity for the following coarse update. On the first step, \(v_0=v_\theta(x_0,t_0)\); for \(k>0\), \(v_k\) is the cached evaluation from the preceding intermediate state \(\tilde x_k\), rather than a fresh evaluation at the corrected state \(x_k\). The same evaluation therefore informs the remaining-residual prediction and seeds the next base step, preserving Euler's backbone-evaluation count. On the final step, the corrector is omitted because its evaluation could not be reused, so the final state is \(\tilde x_{k+1}\).

\subsection{Compact Residual Representation}
\label{sec:prop2}

The predictor and corrector residuals have the same dimensionality as the image or latent sampling state, making direct prediction impractical. TraceSampler therefore makes each residual predictable through a compact decomposition into three components:
\[
	\text{residual error}
	\;\approx\;
	\text{mean residual}
	+
	\text{explicit-direction component}
	+
	\text{low-rank residual component}.
\]

The mean residual captures the correction typically required at a given
sampling step. The explicit-direction component captures variation associated
with the current sampling trace, while the low-rank residual component captures
recurring variation that remains after those directions are accounted for.
The coefficient heads therefore need only predict how strongly the explicit
directions and low-rank basis vectors contribute for the current sample.

The predictor and corrector use the same compact form:
\begin{align}
\hat r_{p,k} &= \bar r_{p,k}+P_k\alpha_{p,k}+D_{p,k}\beta_{p,k},
\label{eq:predictor_residual}\\
\hat r_{c,k} &= \bar r_{c,k}+\tilde C_k\alpha_{c,k}+D_{c,k}\beta_{c,k}.
\label{eq:corrector_residual}
\end{align}

Here, \(\bar r_{s,k}\) is the stage-specific mean residual, \(P_k\) and
\(\tilde C_k\) are matrices whose columns are the predictor and corrector
trajectory directions, respectively, and \(D_{s,k}\) is the fitted low-rank
basis. The vectors \(\alpha_{s,k}\) and \(\beta_{s,k}\) are the
sample-dependent coefficients for the trajectory directions and low-rank
basis, respectively. Each column of \(P_k\) and \(\tilde C_k\) has the same
dimensionality as the image or latent sampling state.

The mean residuals and low-rank bases are fitted offline, while the explicit
trajectory directions are constructed from the current sampling path. The
coefficient values are sample-dependent: different trajectories at the same
sampling step can receive different weights and therefore different
corrections.

\paragraph{Predictor directions.}

The predictor acts before a new backbone evaluation is made, so its explicit
directions use only trajectory information already available at the current
sampling step. Two directions are constructed:
\[
	h_k
	=
	x_k-x_{k-1},
	\qquad
	m_k
	=
	\Delta t_kv_k-h_k.
\]
The direction \(h_k\) is the displacement made by the previous sampling step
and is set to zero on the first step. The direction \(m_k\) measures how the
current coarse displacement differs from the previous displacement. Together,
they describe the recent sampling motion and how that motion is changing. The
directions are normalised before forming the predictor direction matrix
\(P_k\).

\paragraph{Corrector directions.}

The corrector acts after evaluating the backbone at the intermediate state
\(\tilde x_{k+1}\), giving the new velocity \(v_{k+1}\). It can therefore
construct additional directions using both the predictor update and this new
backbone information. Let \(\epsilon_0=x_0\) denote the initial noise. Four directions are used:
\[
\begin{aligned}
s'_k &= \Delta t_k v_{k+1}-(\tilde x_{k+1}-x_k),
&\qquad \delta_k &= \epsilon_0-(\tilde x_{k+1}-t_{k+1}v_{k+1}),\\
q_k &= \tilde x_{k+1}-x_{k+1}^{\mathrm{base}},
&\qquad \kappa_k &= \Delta t_k(v_k-v_{k-1}).
\end{aligned}
\]
The direction \(s'_k\) compares the displacement suggested by the newly
evaluated velocity with the displacement that produced the intermediate
state. The direction \(\delta_k\) measures the difference between the initial
noise \(\epsilon_0\) and the noise-state estimate implied by the intermediate
state and newly evaluated velocity. The direction \(q_k\) is the predictor
correction itself, allowing the corrector to strengthen or partially reverse
it. The direction \(\kappa_k\) captures the change in velocity and is
set to zero on the first step.

Because the four corrector directions can contain overlapping information,
the raw directions are Gram--Schmidt orthogonalised in the fixed order
\[
    [\,s'_k,\delta_k,q_k,\kappa_k\,].
\]
The resulting non-zero directions are then normalised before forming
\(\tilde C_k\).

\subsection{Sample-Dependent Coefficient Prediction}
\label{sec:prop3}

The residual representation defines the available correction components, but
their contribution varies across trajectories. Shallow per-step MLP heads
therefore predict the corresponding coefficients from trajectory features:
\[
(\alpha_{p,k},\beta_{p,k}) = g_{p,k}(\phi_{p,k}),
\qquad
(\alpha_{c,k},\beta_{c,k}) = g_{c,k}(\phi_{c,k}).
\]
Here, \(\phi_{p,k}\) and \(\phi_{c,k}\) are compact feature vectors containing
information available to the predictor and corrector, respectively.

Rather than predicting a full image- or latent-dimensional residual, each head
predicts only a compact set of sample-dependent coefficients for the structured
representation.

Each feature vector concatenates trajectory summaries, projections of the
current state and recent displacement onto the fitted residual basis, the
preceding coefficient-head output from the same stage, and compressed hidden
activations from selected blocks of the frozen backbone. The hidden activations
are taken from backbone evaluations already required by the sampling chain, so
feature construction introduces no additional backbone evaluations. All
features are normalised using statistics from the fitting data. Exact feature
definitions, targeted blocks, compression operations, and input dimensions are
given in Appendix~\ref{app:feature_details}.

\subsection{Offline Fitting}
\label{subsec:residual_representation_fitting}

Offline fitting estimates the shared mean residuals and low-rank bases, constructs sample-specific coefficient targets, and trains the predictor and corrector heads. Fitting follows the deployed chain rather than treating steps independently: at each non-final step the predictor is fitted and applied before corrector targets are formed, and corrected states are carried forward before fitting the next step. Thus the heads are trained on the chained states encountered at inference.

To describe the fitting procedure compactly, let
\[
	M_{p,k}=P_k,
	\qquad
	M_{c,k}=\tilde C_k,
\]
and let \(s\in\{p,c\}\) denote the correction stage.

For each deployed correction stage, fitting proceeds in three parts. First, the
mean residual is estimated. Second, the weights of the explicit trajectory
directions are fitted by least squares. Finally, PCA is applied to the
variation that remains, giving the low-rank basis and its coefficient targets.

Suppose the fitting set contains \(N\) trajectories, indexed by
\(i\in\{1,\ldots,N\}\). Let \(r_{s,k}^{(i)}\) denote the residual target for
trajectory \(i\) at step \(k\) and stage \(s\), and let
\(M_{s,k}^{(i)}\) denote its explicit direction matrix.

The mean residual, least-squares direction weights, and remaining variation are
\[
\begin{aligned}
\bar r_{s,k} &= \frac{1}{N}\sum_{i=1}^{N}r_{s,k}^{(i)},\\
\alpha_{s,k}^{(i)} &= \arg\min_{\alpha}\left\|r_{s,k}^{(i)}-\bar r_{s,k}-M_{s,k}^{(i)}\alpha\right\|_2^2,\\
\eta_{s,k}^{(i)} &= r_{s,k}^{(i)}-\bar r_{s,k}-M_{s,k}^{(i)}\alpha_{s,k}^{(i)}.
\end{aligned}
\]

A rank-\(K\) PCA basis is fitted to the remainder and projected to obtain the
low-rank weights, giving
\[
\begin{aligned}
D_{s,k} &= \mathrm{PCA}_{K}\!\left(\{\eta_{s,k}^{(i)}\}_{i=1}^{N}\right),\\
\beta_{s,k}^{(i)} &= D_{s,k}^{\top}\eta_{s,k}^{(i)},\\
r_{s,k}^{(i)} &\approx \bar r_{s,k}+M_{s,k}^{(i)}\alpha_{s,k}^{(i)}+D_{s,k}\beta_{s,k}^{(i)}.
\end{aligned}
\]

Since the reference residual is known during offline fitting, the coefficient
vectors \(\alpha_{s,k}^{(i)}\) and \(\beta_{s,k}^{(i)}\) can be calculated
directly. They become the sample-specific training targets for the coefficient
head \(g_{s,k}\):
\[
	g_{s,k}\!\left(\phi_{s,k}^{(i)}\right)
	\approx
	\left(
		\alpha_{s,k}^{(i)},
		\beta_{s,k}^{(i)}
	\right).
\]

A separate predictor head is fitted for every sampling step, and a separate
corrector head is fitted for every non-final sampling step. Each head is a
single-hidden-layer MLP whose final layer is initialised to output the mean
coefficient target before training, providing a shared starting point for
trajectory-specific refinement.
Implementation details are provided in Appendix~\ref{app:head_training}.

At inference, the reference residual is no longer needed. The saved mean
residuals and low-rank bases are loaded, the explicit directions are
constructed from the current trajectory, and the coefficient heads predict the
sample-dependent coefficients. These quantities are combined using
Eqs.~\eqref{eq:predictor_residual} and \eqref{eq:corrector_residual} to
reconstruct the predictor and corrector corrections.

\begin{figure}[!htbp]
    \centering
    \includegraphics[
        width=\linewidth,
        height=\textheight,
        keepaspectratio
    ]{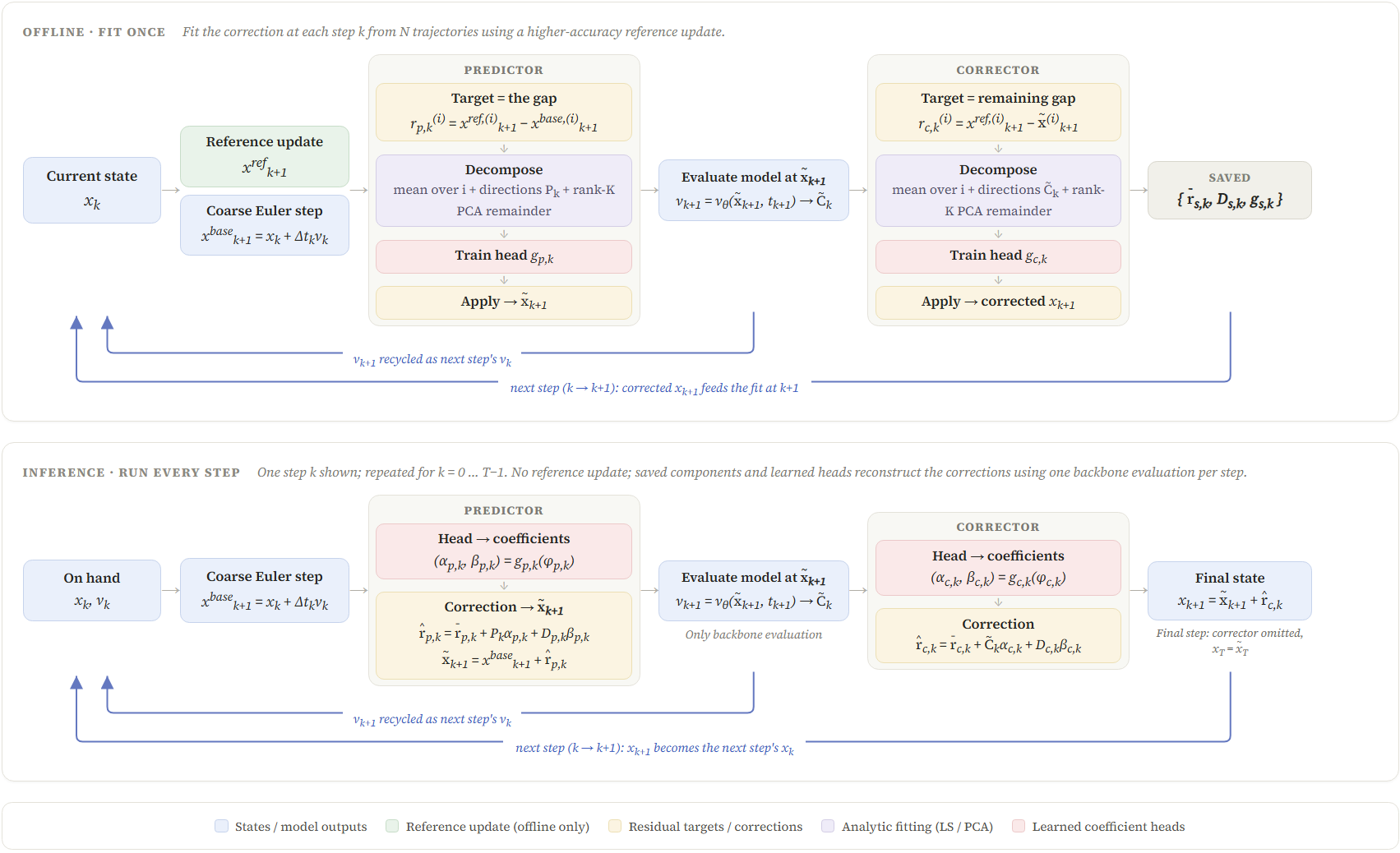}
    \caption{Offline fitting and inference workflow of TraceSampler.}
    \label{fig:proposed_framework_final}
\end{figure}

\paragraph{Complete workflow.} TraceSampler predicts coefficients rather than full residual tensors, separating offline-fitted correction geometry from sample-specific adaptation. Figure~\ref{fig:proposed_framework_final} summarises the pipeline: offline fitting constructs residual supervision and trains the heads; inference uses the saved representation with the frozen backbone, carrying each corrected state into the next step.

\FloatBarrier

\subsection{Optional Timestep-Schedule Optimisation}
\label{sec:time_schedule_optimisation}

TraceSampler can optionally select a low-step schedule offline before fitting the final residual representation and heads. A probe propagates each step's unrepresented residual through the remaining base-sampler rollout. Their Gram matrix $G$ captures individual endpoint-error contributions and cross-step interactions. Rescaling the probe step sizes gives the objective ($p=2$):
\begin{equation}
E_{\mathrm{sched}}(\Delta t)
=\sum_{i,j}G_{ij}
\left(\frac{\Delta t_i}{\Delta t_i^{\mathrm{probe}}}\right)^p
\left(\frac{\Delta t_j}{\Delta t_j^{\mathrm{probe}}}\right)^p,
\qquad p=2.
\label{eq:schedule_objective}
\end{equation}
Proposed schedules are remeasured on matched probe trajectories and accepted only when propagated endpoint-error MSE improves by more than twice the combined standard error. The selected schedule is fixed for fitting and inference; Appendix~\ref{app:schedule_analysis} gives full details and representative schedules.

\section{Experiments}
\label{sec:experiments}

\subsection{Experimental Setup}
\label{sec:experimental_setup}

We evaluate TraceSampler with Rectified Flow on CIFAR-10
\citep{liu2022flow,krizhevsky2009learning}, SiT-XL/2 on ImageNet-256
\citep{ma2024sit,deng2009imagenet}, and SD3.5 Medium at \(512\times512\)
\citep{esser2024scaling}. CIFAR-10 supports controlled ablations, ImageNet-256
tests scaling, and SD3.5 evaluates the text-conditioned setting. We report FID,
IS \citep{heusel2017gans,salimans2016improved}, and trajectory MSE on CIFAR-10
and ImageNet-256; SD3.5 additionally uses ImageReward (IR), CLIP similarity,
GenEval, and HPSv2.1 
\citep{xu2023imagereward,radford2021learning,ghosh2023geneval,wu2023human}.

Trajectory MSE measures deviation from the higher-step reference at matched timesteps, with reductions relative to Euler. Residual representations use all \(24{,}576\) trajectories, while coefficient heads use a \(19{,}661/4{,}915\) train/validation split. Table~\ref{tab:main_parameters} summarises the main configurations; protocols and reference-substep details are in Appendix~\ref{app:baseline_configurations}, with hardware in Appendix~\ref{app:preparation_measurements}.

\begin{table}[!htbp]
    \centering
    \scriptsize
    \setlength{\tabcolsep}{4.2pt}
    \renewcommand{\arraystretch}{0.96}
    \caption{Main TraceSampler fitting and sampling configurations.}
    \label{tab:main_parameters}
    \begin{tabular*}{\linewidth}{@{\extracolsep{\fill}}lcccc@{}}
        \toprule
        Model/config. & Offline reference & PCA rank & MLP width & Schedule \\
        \midrule
        RF CIFAR-10 & 120-step Heun & $K=32$ & $h=12288$ & Optimised \\
        SiT-XL/2 & 120-step Heun & $K=32$ & $h=12288$ & Optimised \\
        SD3.5 Medium (8-step) & 64-step Euler & $K=32$ & $h=24576$ & Shift $=3$ \\
        SD3.5 Medium (10-step) & 60-step Euler & $K=32$ & $h=24576$ & Shift $=3$ \\
        \bottomrule
    \end{tabular*}
\end{table}

\subsection{Results}
\label{sec:results}

The results first isolate TraceSampler's design choices on CIFAR-10, then evaluate scaling to ImageNet-256 and transfer to SD3.5 Medium, before examining computational cost.

% ============================================================
\subsection{Controlled Ablations}
\label{sec:rf_ablations}

Controlled CIFAR-10 experiments isolate the main design choices. \textbf{Component build-up.} Table~\ref{tab:rf_buildup} shows the incremental construction from the same 8-step Euler sampler: predictor directions, corrector directions, the low-rank PCA basis, and timestep optimisation.

\begin{table}[!htbp]
    \centering
    \scriptsize
    \setlength{\tabcolsep}{6pt}
    \renewcommand{\arraystretch}{0.96}
    \caption{Component build-up at 8 steps.}
    \label{tab:rf_buildup}
    \begin{tabular}{lccccc}
        \toprule
        Method
         & Schedule
         & \(K\)
         & FID \(\downarrow\)
         & IS \(\uparrow\)
         & Trajectory MSE reduction \(\uparrow\) \\
        \midrule
        Base Euler
         & Uniform
         & 0
         & 15.729
         & 6.321
         & 0.0\% \\
        Predictor directions only
         & Uniform
         & 0
         & 9.705
         & 7.757
         & 34.7\% \\
        Predictor + corrector directions
         & Uniform
         & 0
         & 4.101
         & 8.612
         & 80.5\% \\
        Full TraceSampler
         & Uniform
         & 32
         & 3.666
         & 8.626
         & 87.5\% \\
        Full TraceSampler + schedule
         & Optimised
         & 32
         & \textbf{2.798}
         & 8.552
         & \textbf{90.8\%} \\
        \bottomrule
    \end{tabular}
\end{table}

The explicit directions drive most of the gain: the corrector gives the largest jump, reaching FID \(4.101\) and an \(80.5\%\) trajectory MSE reduction after observing the post-predictor backbone. PCA improves FID to \(3.666\), while schedule optimisation reaches \(2.798\) and \(90.8\%\). Thus trajectory-aligned structure dominates, the low-rank basis captures the remainder, and schedule optimisation adds a smaller final gain.

\textbf{Sample-dependent prediction.}
To isolate the learned heads, we replace them with fixed step- and stage-specific mean oracle coefficients while keeping the residual bases, sample-dependent directions, schedule, and sampling configuration unchanged.

\begin{figure}[!htbp]
    \centering
    \includegraphics[width=0.76\linewidth]{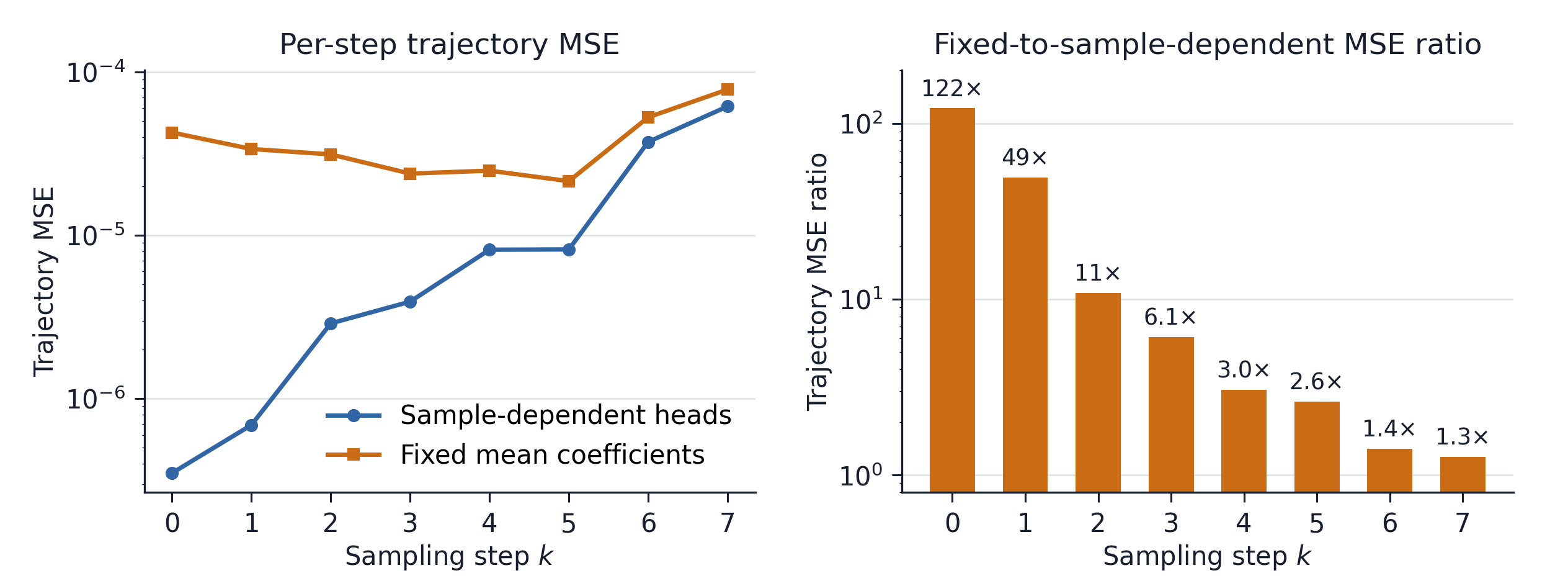}
    \caption{Per-step trajectory MSE and the fixed-to-sample-dependent error
        ratio.}
    \label{fig:rf_sample_dependence}
\end{figure}

\begin{table}[!htbp]
    \centering
    \scriptsize
    \setlength{\tabcolsep}{8pt}
    \renewcommand{\arraystretch}{0.96}
    \caption{Coefficient-prediction ablation at 8 steps.}
    \label{tab:rf_sample_dependence}
    \begin{tabular}{lrrr}
        \toprule
        Coefficient prediction
        & FID \(\downarrow\)
        & IS \(\uparrow\)
        & Trajectory MSE reduction \(\uparrow\) \\
        \midrule
        Fixed mean coefficients
        & 3.240
        & \textbf{8.563}
        & 81.5\% \\
        Sample-dependent heads
        & \textbf{2.798}
        & 8.552
        & \textbf{90.8\%} \\
        \bottomrule
    \end{tabular}
\end{table}

The fixed-coefficient variant already reduces trajectory MSE by \(81.5\%\),
showing that coarse-step error contains a large systematic component captured by
the shared representation. Making the coefficients trace-dependent increases
the reduction to \(90.8\%\) and improves FID from \(3.240\) to \(2.798\).
Thus the gain is not only a better average correction: conditioning on the
current trajectory explains additional residual variation that a step-wise
fixed correction cannot. Figure~\ref{fig:rf_sample_dependence} shows that this
sample dependence is strongest at early coarse steps. Further sweeps are in
Appendix~\ref{app:rf_design_ablations}.

% ============================================================
\subsection{Matched-Budget Sampling on CIFAR-10 and ImageNet-256}
\label{sec:matched_budget_results}

\begin{table}[!htbp]
    \centering
    \scriptsize
    \setlength{\tabcolsep}{1.55pt}
    \renewcommand{\arraystretch}{0.96}
    \caption{CIFAR-10 matched-budget comparison.}
    \label{tab:rf_sampler_comparison}
    \begin{tabular*}{\linewidth}{@{\extracolsep{\fill}}lrrrrrrrrr@{}}
        \toprule
        \multirow{2}{*}{Budget}
         & \multicolumn{2}{c}{Euler}
         & \multicolumn{2}{c}{IIA--Euler}
         & \multicolumn{2}{c}{UniPC}
         & \multicolumn{2}{c}{TraceSampler}
         & \multirow{2}{*}{\shortstack{Traj. MSE\\red. \(\uparrow\)}} \\
        \cmidrule(lr){2-3}
        \cmidrule(lr){4-5}
        \cmidrule(lr){6-7}
        \cmidrule(lr){8-9}
         & FID \(\downarrow\)
         & IS \(\uparrow\)
         & FID \(\downarrow\)
         & IS \(\uparrow\)
         & FID \(\downarrow\)
         & IS \(\uparrow\)
         & FID \(\downarrow\)
         & IS \(\uparrow\)
         & \\
        \midrule
        5 steps  & 33.149 & 4.566 & 22.253 & 5.783 & 12.817 & 7.238 & \textbf{4.373} & \textbf{8.239} & \textbf{89.9\%} \\
        6 steps  & 24.143 & 5.327 & 14.940 & 6.510 & 9.333 & 7.730 & \textbf{3.560} & \textbf{8.344} & \textbf{88.8\%} \\
        8 steps  & 15.729 & 6.321 & 8.846 & 7.357 & 7.751 & 8.011 & \textbf{2.798} & \textbf{8.552} & \textbf{90.8\%} \\
        10 steps & 11.782 & 6.886 & 6.668 & 7.670 & 6.429 & 8.169 & \textbf{2.692} & \textbf{8.556} & \textbf{90.5\%} \\
        12 steps & 9.632 & 7.205 & 5.671 & 7.850 & 5.778 & 8.256 & \textbf{2.641} & \textbf{8.517} & \textbf{89.5\%} \\
        \midrule
        Heun 120-step reference
         & \multicolumn{9}{c}{FID \(2.601\),\quad IS \(8.570\)} \\
        \bottomrule
    \end{tabular*}
\end{table}
\FloatBarrier

TraceSampler leads FID and IS at every CIFAR-10 budget, with the largest
absolute gains when integration is coarsest: at 5 steps, FID drops from
\(33.149\) for Euler and \(12.817\) for UniPC to \(4.373\). By 10 steps,
FID reaches \(2.692\), close to the 120-step Heun value \(2.601\), together
with a \(90.5\%\) trajectory-MSE reduction. The shrinking gap as the budget
grows is expected if TraceSampler is primarily removing coarse-integration
error rather than replacing the underlying generative model.

\begin{table}[!htbp]
    \centering
    \scriptsize
    \setlength{\tabcolsep}{1.55pt}
    \renewcommand{\arraystretch}{1.02}
    \caption{ImageNet-256 matched-budget comparison.}
    \label{tab:sit_sampler_comparison}
    \begin{tabular*}{\linewidth}{@{\extracolsep{\fill}}lrrrrrrrrr@{}}
        \toprule
        \multirow{2}{*}{Budget}
         & \multicolumn{2}{c}{Euler}
         & \multicolumn{2}{c}{IIA--Euler}
         & \multicolumn{2}{c}{UniPC}
         & \multicolumn{2}{c}{TraceSampler}
         & \multirow{2}{*}{\shortstack{Traj. MSE\\red. \(\uparrow\)}} \\
        \cmidrule(lr){2-3}
        \cmidrule(lr){4-5}
        \cmidrule(lr){6-7}
        \cmidrule(lr){8-9}
         & FID \(\downarrow\)
         & IS \(\uparrow\)
         & FID \(\downarrow\)
         & IS \(\uparrow\)
         & FID \(\downarrow\)
         & IS \(\uparrow\)
         & FID \(\downarrow\)
         & IS \(\uparrow\)
         & \\
        \midrule
        6 steps
         & 19.786 & 114.66 & 13.233 & 151.307 & 4.630 & 206.75 & \textbf{3.624} & \textbf{209.74} & \textbf{79.8\%} \\
        8 steps
         & 10.675 & 161.18 & 7.405 & 185.997 & 3.105 & 227.55 & \textbf{2.418} & \textbf{235.21} & \textbf{83.2\%} \\
        10 steps
         & 7.189 & 186.80 & 5.204 & 204.654 & 2.544 & 238.45 & \textbf{2.189} & \textbf{245.15} & \textbf{85.6\%} \\
        \midrule
        Heun 120-step reference
         & \multicolumn{9}{c}{FID \(2.056\),\quad IS \(254.16\)} \\
        \bottomrule
    \end{tabular*}
\end{table}

The same pattern holds after scaling to ImageNet-256. TraceSampler improves
both FID and IS at every tested budget, reducing 6-step FID from \(19.786\)
for Euler and \(4.630\) for UniPC to \(3.624\). At 10 steps it reaches FID
\(2.189\), IS \(245.15\), and an \(85.6\%\) trajectory-MSE reduction,
approaching the 120-step reference. The agreement between trajectory error and
generation metrics across two different backbones supports endpoint-residual
prediction as a transferable correction signal; qualitative results are in
Appendix~\ref{app:sit_qualitative}.

% ============================================================
\subsection{SD3.5 Medium Text-to-Image Generation}
\label{sec:sd35_results}

We test text-conditioned transfer on SD3.5 Medium at matched 8- and 10-step budgets, jointly measuring quality, prompt following, preference, and trajectory matching.

\begin{table}[!htbp]
    \centering
    \scriptsize
    \setlength{\tabcolsep}{1.9pt}
    \renewcommand{\arraystretch}{0.96}
    \caption{SD3.5 Medium results.}
    \label{tab:sd35_quality}
    \begin{tabular*}{\linewidth}{@{\extracolsep{\fill}}lrrrrrrr@{}}
        \toprule
        Sampler & IR \(\uparrow\) & FID \(\downarrow\) & IS \(\uparrow\) & CLIP \(\uparrow\) & GenEval \(\uparrow\) & HPS \(\uparrow\) & \shortstack{Traj. MSE\\red. \(\uparrow\)} \\
        \midrule
        Euler 8 & 0.611 & 17.320 & 39.23 & 0.2698 & 0.60997 & 0.2572 & -- \\
        Euler 10 & 0.752 & 17.368 & 40.48 & 0.2697 & 0.64152 & 0.2702 & -- \\
        \addlinespace
        TraceSampler 8 & \textbf{0.862} & \textbf{14.616} & \textbf{43.28} & 0.2661 & \textbf{0.64554} & \textbf{0.2807} & \textbf{67.9\%} \\
        TraceSampler 10 & \textbf{0.906} & \textbf{15.478} & \textbf{43.05} & 0.2659 & \textbf{0.66811} & \textbf{0.2878} & \textbf{70.7\%} \\
        \midrule
        Euler 64 eval. ref. & 0.973 & 18.059 & 43.95 & 0.2651 & 0.68275 & 0.2968 & -- \\
        \bottomrule
    \end{tabular*}
\end{table}
\FloatBarrier

TraceSampler reduces trajectory MSE by \(67.9\%\) at 8 steps and
\(70.7\%\) at 10 while improving ImageReward, FID, IS, GenEval, and HPSv2.1
at both budgets. At 10 steps, ImageReward rises from \(0.752\) to \(0.906\),
GenEval from \(0.64152\) to \(0.66811\), and HPSv2.1 from \(0.2702\) to
\(0.2878\). CLIP decreases slightly, and the 64-evaluation reference itself
does not dominate every distributional metric (e.g., FID), so closeness to the
reference should not be read as monotonic improvement on every metric. The
consistent gains in preference and compositional metrics nevertheless show
that residual correction transfers beyond unconditional and class-conditional
settings.

Figure~\ref{fig:sd35_qualitative} gives a matched-noise comparison with Euler and the
64-step reference; prompts are in Appendix~\ref{app:sd35_qualitative_prompts}.
TraceSampler recovers detail and structure lost by 8-step Euler while remaining
visually close to the 64-step reference, consistent with Table~\ref{tab:sd35_quality}.

\begin{figure}[H]
    \centering
    \includegraphics[width=0.89\linewidth]{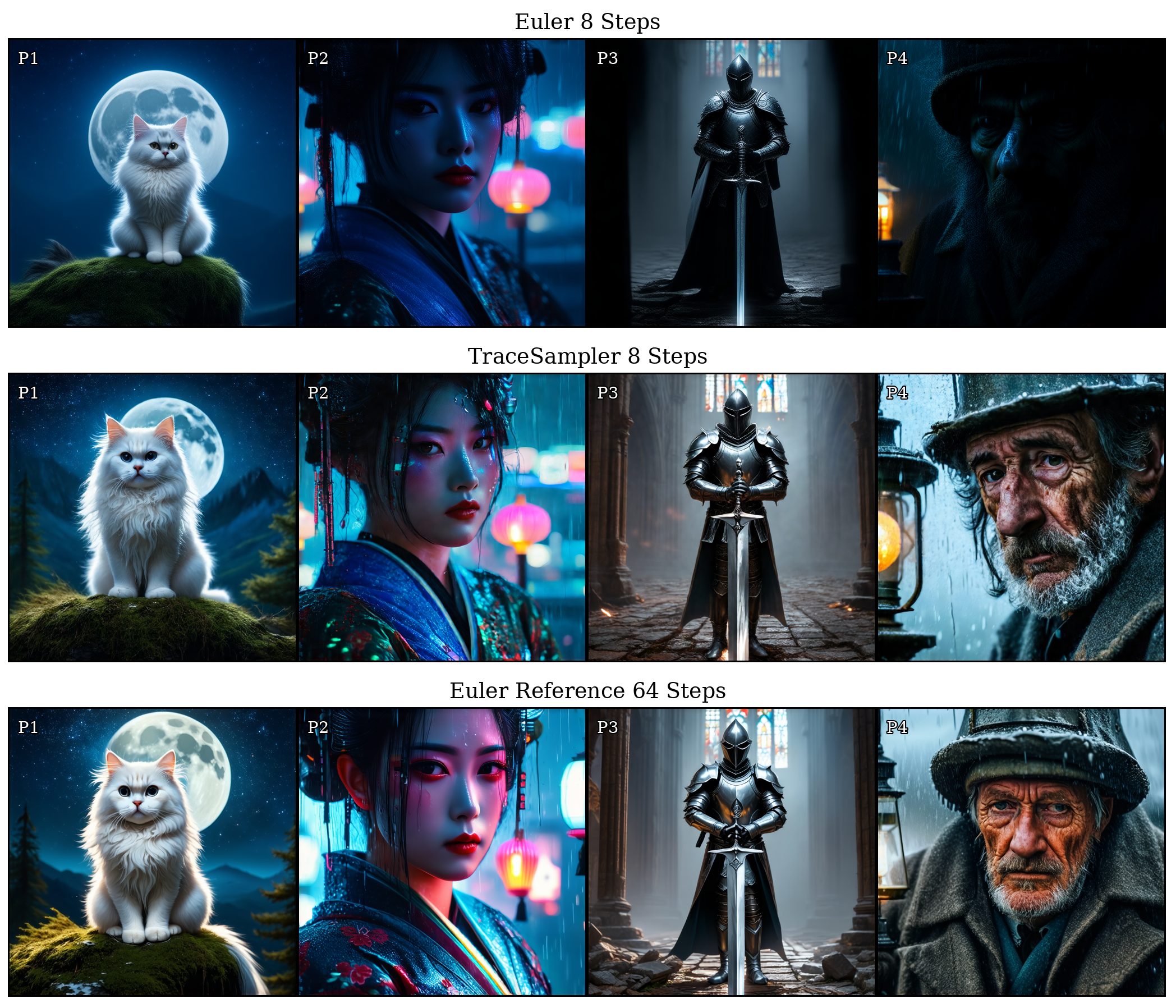}
    \caption{SD3.5 Medium qualitative comparison at matched noise.}
    \label{fig:sd35_qualitative}
\end{figure}

% ============================================================
\subsection{Computational Cost}
\label{sec:computational_cost}

Reusing each intermediate-state backbone evaluation in the following base update preserves Euler's backbone-evaluation budget; Table~\ref{tab:inference_cost} isolates the added compute, throughput, and memory cost.

\begin{table}[H]
    \centering
    \scriptsize
    \setlength{\tabcolsep}{1.4pt}
    \renewcommand{\arraystretch}{0.96}
    \caption{Inference cost at 8 steps; FLOPs are per step.}
    \label{tab:inference_cost}
    \begin{tabular*}{\linewidth}{@{\extracolsep{\fill}}lrrrrrrr@{}}
        \toprule
        Model
        & \multicolumn{3}{c}{GFLOPs/step}
        & \multicolumn{2}{c}{Throughput (images/s)}
        & \multicolumn{2}{c}{Peak VRAM (GiB)} \\
        \cmidrule(lr){2-4}
        \cmidrule(lr){5-6}
        \cmidrule(lr){7-8}
        & Euler & Trace & Increase & Euler & Trace & Euler & Trace \\
        \midrule
        Rectified Flow & 21.6931 & 21.7714 & 0.3610\% & 321.45 & 316.05 & 6.77 & 7.52 \\
        SiT-XL/2 & 474.4857 & 475.0351 & 0.1158\% & 31.07 & 30.97 & 7.92 & 8.51 \\
        SD3.5 Medium & 4275.6574 & 4278.4379 & 0.0650\% & 4.144 & 4.126 & 11.60 & 14.22 \\
        \bottomrule
    \end{tabular*}
\end{table}

Added compute is only \(0.0650\%\)--\(0.3610\%\), and measured throughput remains within \(1.7\%\) of Euler across all backbones. The main deployment trade-off is memory: peak VRAM rises by \(0.59\)--\(2.62\) GiB, mainly from the additional parameters of the coefficient heads. Changing the step budget or schedule requires one-time refitting; head sizes and offline costs are in Appendix~\ref{app:additional_results}.

\section{Conclusion}
\label{sec:conclusion}

TraceSampler models coarse-step endpoint error as a structured, sample-dependent prediction target and corrects it without extra backbone evaluations. At 10 steps, it reaches FID $2.692$ on CIFAR-10 and $2.189$ on ImageNet-256 with $90.5\%$ and $85.6\%$ trajectory MSE reductions; on SD3.5 Medium, ImageReward rises from $0.752$ to $0.906$ as trajectory MSE falls $70.7\%$. Across all three settings, endpoint error contains predictable structure tied to the sampling trace that can be exploited for correction. These gains add less than $0.4\%$ inference FLOPs with negligible wall-clock overhead.

\bibliography{mybib}

@inproceedings{deng2009imagenet,
  title={{ImageNet}: A Large-Scale Hierarchical Image Database},
  author={Deng, Jia and Dong, Wei and Socher, Richard and Li, Li-Jia and Li, Kai and Fei-Fei, Li},
  booktitle={2009 IEEE Conference on Computer Vision and Pattern Recognition},
  pages={248--255},
  year={2009}
}

@inproceedings{esser2024scaling,
  title={Scaling Rectified Flow Transformers for High-Resolution Image Synthesis},
  author={Esser, Patrick and Kulal, Sumith and Blattmann, Andreas and Entezari, Rahim and M{\"u}ller, Jonas and Saini, Harry and Levi, Yam and Lorenz, Dominik and Sauer, Axel and Boesel, Frederic and others},
  booktitle={Forty-first International Conference on Machine Learning},
  year={2024}
}

@inproceedings{ghosh2023geneval,
  title={{GenEval}: An Object-Focused Framework for Evaluating Text-to-Image Alignment},
  author={Ghosh, Dhruba and Hajishirzi, Hannaneh and Schmidt, Ludwig},
  booktitle={Advances in Neural Information Processing Systems},
  volume={36},
  pages={52132--52152},
  year={2023}
}

@inproceedings{heusel2017gans,
  title={{GANs} Trained by a Two Time-Scale Update Rule Converge to a Local Nash Equilibrium},
  author={Heusel, Martin and Ramsauer, Hubert and Unterthiner, Thomas and Nessler, Bernhard and Hochreiter, Sepp},
  booktitle={Advances in Neural Information Processing Systems},
  volume={30},
  year={2017}
}

@techreport{krizhevsky2009learning,
  title={Learning Multiple Layers of Features from Tiny Images},
  author={Krizhevsky, Alex and Hinton, Geoffrey},
  year={2009}
}

@inproceedings{lin2014microsoft,
  title={Microsoft {COCO}: Common Objects in Context},
  author={Lin, Tsung-Yi and Maire, Michael and Belongie, Serge and Hays, James and Perona, Pietro and Ramanan, Deva and Doll{\'a}r, Piotr and Zitnick, C. Lawrence},
  booktitle={European Conference on Computer Vision},
  pages={740--755},
  year={2014}
}

@inproceedings{lipman2022flow,
  title={Flow Matching for Generative Modeling},
  author={Lipman, Yaron and Chen, Ricky T. Q. and Ben-Hamu, Heli and Nickel, Maximilian and Le, Matthew},
  booktitle={International Conference on Learning Representations},
  year={2023},
}

@inproceedings{liu2022flow,
  title={Flow Straight and Fast: Learning to Generate and Transfer Data with Rectified Flow},
  author={Liu, Xingchao and Gong, Chengyue and Liu, Qiang},
  booktitle={International Conference on Learning Representations},
  year={2023},
}

@article{lu2025dpm,
  title={{DPM-Solver++}: Fast Solver for Guided Sampling of Diffusion Probabilistic Models},
  author={Lu, Cheng and Zhou, Yuhao and Bao, Fan and Chen, Jianfei and Li, Chongxuan and Zhu, Jun},
  journal={Machine Intelligence Research},
  volume={22},
  number={4},
  pages={730--751},
  year={2025}
}

@inproceedings{ma2024sit,
  title={{SiT}: Exploring Flow and Diffusion-Based Generative Models with Scalable Interpolant Transformers},
  author={Ma, Nanye and Goldstein, Mark and Albergo, Michael S. and Boffi, Nicholas M. and Vanden-Eijnden, Eric and Xie, Saining},
  booktitle={European Conference on Computer Vision},
  pages={23--40},
  year={2024}
}

@inproceedings{radford2021learning,
  title={Learning Transferable Visual Models from Natural Language Supervision},
  author={Radford, Alec and Kim, Jong Wook and Hallacy, Chris and Ramesh, Aditya and Goh, Gabriel and Agarwal, Sandhini and Sastry, Girish and Askell, Amanda and Mishkin, Pamela and Clark, Jack and others},
  booktitle={International Conference on Machine Learning},
  pages={8748--8763},
  year={2021}
}

@inproceedings{salimans2016improved,
  title={Improved Techniques for Training {GANs}},
  author={Salimans, Tim and Goodfellow, Ian and Zaremba, Wojciech and Cheung, Vicki and Radford, Alec and Chen, Xi},
  booktitle={Advances in Neural Information Processing Systems},
  volume={29},
  year={2016}
}

@inproceedings{song2020score,
  title={Score-Based Generative Modeling through Stochastic Differential Equations},
  author={Song, Yang and Sohl-Dickstein, Jascha and Kingma, Diederik P. and Kumar, Abhishek and Ermon, Stefano and Poole, Ben},
  booktitle={International Conference on Learning Representations},
  year={2021},
}

@inproceedings{wang2025pas,
  title={Diffusion Sampling Correction via Approximately 10 Parameters},
  author={Wang, Guangyi and Peng, Wei and Li, Lijiang and Chen, Wenyu and Cai, Yuren and Su, Song-Zhi},
  booktitle={Proceedings of the 42nd International Conference on Machine Learning},
  series={Proceedings of Machine Learning Research},
  volume={267},
  pages={64023--64043},
  year={2025}
}

@inproceedings{wang2025differentiable,
  title={Differentiable Solver Search for Fast Diffusion Sampling},
  author={Wang, Shuai and Li, Zexian and Zhang, Qipeng and Song, Tianhui and Li, Xubin and Ge, Tiezheng and Zheng, Bo and Wang, Limin},
  booktitle={Proceedings of the 42nd International Conference on Machine Learning},
  series={Proceedings of Machine Learning Research},
  volume={267},
  pages={63549--63568},
  year={2025}
}

@article{wu2023human,
  title={Human Preference Score v2: A Solid Benchmark for Evaluating Human Preferences of Text-to-Image Synthesis},
  author={Wu, Xiaoshi and Hao, Yiming and Sun, Keqiang and Chen, Yixiong and Zhu, Feng and Zhao, Rui and Li, Hongsheng},
  journal={arXiv preprint arXiv:2306.09341},
  year={2023}
}

@inproceedings{xu2023imagereward,
  title={{ImageReward}: Learning and Evaluating Human Preferences for Text-to-Image Generation},
  author={Xu, Jiazheng and Liu, Xiao and Wu, Yuchen and Tong, Yuxuan and Li, Qinkai and Ding, Ming and Tang, Jie and Dong, Yuxiao},
  booktitle={Advances in Neural Information Processing Systems},
  volume={36},
  pages={15903--15935},
  year={2023}
}

@inproceedings{zhang2023accelerating,
  title={On Accelerating Diffusion-Based Sampling Processes via Improved Integration Approximation},
  author={Zhang, Guoqiang and Niwa, Kenta and Kleijn, W. Bastiaan},
  booktitle={International Conference on Learning Representations},
  year={2024}
}

@inproceedings{zhao2023unipc,
  title={{UniPC}: A Unified Predictor-Corrector Framework for Fast Sampling of Diffusion Models},
  author={Zhao, Wenliang and Bai, Lujia and Rao, Yongming and Zhou, Jie and Lu, Jiwen},
  booktitle={Advances in Neural Information Processing Systems},
  volume={36},
  pages={49842--49869},
  year={2023}
}

@inproceedings{zhao2024dcsolver,
  title={{DC-Solver}: Improving Predictor-Corrector Diffusion Sampler via Dynamic Compensation},
  author={Zhao, Wenliang and Wang, Haolin and Zhou, Jie and Lu, Jiwen},
  booktitle={European Conference on Computer Vision},
  pages={450--466},
  year={2024},
  organization={Springer}
}
\bibliographystyle{tracesampler_references}
\newpage
\appendix
\raggedbottom
\section*{Appendices}

\section{Implementation Details}
\label{app:implementation_details}

\subsection{Feature Construction and Targeted Blocks}
\label{app:feature_details}

Section~\ref{sec:tracesampler} leaves the coefficient-head features
\(\phi_{s,k}\) abstract; this section specifies exactly what is available to
each stage and how those signals are compressed. At sampling step \(k\), the
predictor's state features use the corrected state \(x_k\). Its velocity
statistics and hidden activations use the backbone evaluation already
available at \(t_k\): the evaluation at \(x_0\) for \(k=0\), and the evaluation
computed at the preceding intermediate state \(\tilde x_k\) for \(k>0\). The
corrector uses \(\tilde x_{k+1}\) together with the newly computed velocity and
hidden activations at that state.

\paragraph{Handcrafted trajectory features.}
The compact trajectory-summary block was selected using held-out permutation
importance. Candidate features whose permutation caused negligible degradation
for both coefficient heads, including the state spread, state--velocity cosine,
and global velocity norm, were removed. The retained features are the
per-channel spatial standard deviation of the velocity, the norm of the most
recent displacement, and the displacement--velocity cosine. These provide
complementary information about update scale and changes in the local
trajectory direction.

For the handcrafted trajectory summaries, the displacement is defined
relative to the state associated with the available backbone velocity. For
the predictor,
\[
    d_{p,k}
    =
    \begin{cases}
        0, & k=0,\\
        \tilde x_k-x_{k-1}, & k>0,
    \end{cases}
\]
while for the corrector,
\[
    d_{c,k}
    =
    \tilde x_{k+1}-x_k.
\]
For stage \(s\in\{p,c\}\), the trajectory-summary block is therefore
\[
    \left[
        \operatorname{std}_{h,w}(v),\;
        \lVert d_{s,k}\rVert_2,\;
        \frac{\langle d_{s,k},v\rangle}
             {\lVert d_{s,k}\rVert_2\lVert v\rVert_2}
    \right],
\]
where \(\operatorname{std}_{h,w}(v)\) denotes the vector of per-channel
standard deviations over the spatial dimensions \(h\) and \(w\). The norm and
cosine are set to zero when the required displacement is unavailable or has
zero norm. For the predictor, \(v=v_k\) is the available backbone velocity;
for the corrector, \(v=v_{k+1}\) is evaluated at
\(\tilde x_{k+1}\).

\paragraph{Residual-basis and history features.}
The residual-basis projections use displacement definitions distinct from the
handcrafted trajectory summaries above:
\[
    d^{\mathrm{basis}}_{p,k}=x_k-x_{k-1},
    \qquad
    d^{\mathrm{basis}}_{c,k}=\tilde x_{k+1}-\tilde x_k,
\]
with unavailable displacements set to zero. The feature vector contains the
current stage state and the corresponding basis-projection displacement
projected onto the low-rank residual basis fitted for that step and stage. The
stage state is \(x_k\) for the predictor and \(\tilde x_{k+1}\) for the
corrector. The raw output of the preceding head at the same correction stage
is concatenated to these features:
\[
    u_{p,k}
    =
    \begin{cases}
        0, & k=0,\\
        g_{p,k-1}(\phi_{p,k-1}), & k>0,
    \end{cases}
    \qquad
    u_{c,k}
    =
    \begin{cases}
        0, & k=0,\\
        g_{c,k-1}(\phi_{c,k-1}), & k>0.
    \end{cases}
\]
Each preceding output contains both the explicit-direction coefficients and
the residual-basis coefficients. No timestep scalar is appended.

\begin{table}[!htbp]
    \centering
    \scriptsize
    \setlength{\tabcolsep}{4pt}
    \renewcommand{\arraystretch}{1.15}
    \caption{Model-specific hidden-activation features. Input dimensions are
        shown as predictor/corrector (P/C).}
    \label{tab:feature_definitions}
    \begin{tabular}{@{}p{0.16\linewidth}>{\raggedright\arraybackslash}p{0.36\linewidth}>{\raggedright\arraybackslash}p{0.29\linewidth}p{0.12\linewidth}@{}}
        \toprule
        Model & Targeted blocks & Compression & Input dim. \\
        \midrule
        Rectified Flow CIFAR-10
        & Zero-indexed NCSN++ residual blocks \(4\), \(19\), \(26\), and
          \(39\)
        & Spatial mean and standard deviation for each activation channel; no
          learned activation projection
        & \(1639/1641\) \\

        SiT-XL/2
        & Conditional halves of \texttt{block.0}, \texttt{block.26}, and
          \texttt{block.27}
        & Fixed block-specific PCA projection to 256 dimensions, followed by
          token-wise mean and standard deviation
        & \(1928/1930\) \\

        SD3.5 Medium
        & Conditional image streams of \texttt{block.7}, \texttt{block.14},
          and \texttt{block.20}
        & Fixed block-specific PCA projection to 256 dimensions, followed by
          token-wise mean and standard deviation
        & \(3092/3094\) \\
        \bottomrule
    \end{tabular}
\end{table}

\paragraph{Targeted backbone blocks.}
For SiT-XL/2 and SD3.5 Medium, the targeted blocks were selected using
layer-informativeness probes. For each candidate block, a ridge readout was
fitted from pooled activation statistics to the leading PCA coefficients of
the reference--base residual, and predictive performance was measured using
held-out \(R^2\) across sampling steps.

For Rectified Flow, four NCSN++ residual blocks were chosen to span the
encoder, bottleneck region, and decoder without collecting activations from
the complete U-Net. The exact blocks used for each model are reported in
Table~\ref{tab:feature_definitions}.

\paragraph{Activation compression.}
For SiT-XL/2 and SD3.5 Medium, each targeted activation is projected to 256
dimensions before its token-wise mean and standard deviation are calculated.
The common projection dimension keeps the head inputs compact and consistent
across blocks with different original widths. Each block-specific projection
is fitted from all sampling times of a 4096-sample Euler rollout. Token-mean
activations are centred, and their leading 256 right singular vectors define
the fixed projection used during coefficient-head fitting and inference.

\paragraph{Feature normalisation.}
Each feature dimension is standardised separately for every sampling step and
correction stage using the mean and standard deviation over all 24,576
collected fitting trajectories. These statistics are calculated before the
per-head training--validation split. A feature dimension whose standard
deviation is below \(10^{-6}\) is assigned a normalisation scale of one. The
stored statistics are reused for validation and inference, where all
denominators are additionally clamped below by \(10^{-6}\).

\subsection{Prediction-Head Training}
\label{app:head_training}

A separate predictor head is fitted for every sampling step, and a separate
corrector head is fitted for every non-final sampling step. Heads are fitted
in sampling order. At each non-final step, the predictor is fitted first and
applied to construct the intermediate states used to fit the corrector. The
resulting corrected states form the starting states for the following sampling
step, so offline fitting follows the same chained state updates used during
inference.

The training target is the concatenation of the oracle trajectory-direction
coefficients and low-rank residual-basis coefficients. The implementation uses
an unweighted mean-squared error over all scalar outputs:
\[
    \mathcal{L}_{s,k}
    =
    \frac{1}{m_{s,k}}
    \left\|
        g_{s,k}(\phi_{s,k})-y_{s,k}
    \right\|_2^2,
\]
where \(y_{s,k}\) is the concatenated oracle coefficient target and
\(m_{s,k}\) is its dimension. The coefficient targets are not standardised;
only the input features are standardised.

\begin{table}[!htbp]
    \centering
    \scriptsize
    \setlength{\tabcolsep}{6pt}
    \renewcommand{\arraystretch}{1.12}
    \caption{Prediction-head architecture and training settings.}
    \label{tab:head_training_settings}
    \begin{tabular}{@{}p{0.27\linewidth}p{0.67\linewidth}@{}}
        \toprule
        Setting & Configuration \\
        \midrule
        Architecture
        & Independent one-hidden-layer ReLU MLP for each predictor step and
          each non-final corrector step; hidden width \(12288\) for Rectified
          Flow and SiT-XL/2 and \(24576\) for SD3.5 Medium \\

        Optimisation
        & AdamW; initial learning rate \(10^{-3}\), weight decay \(10^{-3}\),
          dropout \(0.1\), batch size \(2048\), and 20-epoch linear warm-up \\

        Learning-rate schedule
        & \texttt{ReduceLROnPlateau} with factor \(0.5\) and patience 50 \\

        Stopping and selection
        & Up to \(20{,}000\) epochs;
          early stopping with patience 400 on the \(4{,}915\)-trajectory
          validation set; best validation checkpoint retained \\

        Numerical precision
        & Backbone evaluations in bfloat16 autocast; residual decomposition,
          coefficient targets, feature normalisation, and MLP fitting in
          float32 \\
        \bottomrule
    \end{tabular}
\end{table}

The output layer is initialised with zero weights and a bias equal to the mean
training coefficient vector. Numerical handling of direction normalisation and
corrector Gram--Schmidt orthogonalisation is described in
Appendix~\ref{app:channel_decomposition}. The per-sample coefficient
least-squares systems use isotropic ridge regularisation of \(10^{-8}\).

\subsection{Residual Decomposition Across Channels}
\label{app:channel_decomposition}

A separate residual decomposition is fitted for each sampling step and
correction stage using all collected fitting trajectories. After subtracting
the stage mean, the explicit-direction coefficients are fitted by the
ridge-regularised least-squares system described in
Section~\ref{subsec:residual_representation_fitting}, and PCA is applied to
the remaining component \(\eta_{s,k}^{(i)}\). Near-zero directions are
treated as zero during normalisation and Gram--Schmidt orthogonalisation for
numerical stability.

Each explicit trajectory direction spans the full sampling state and therefore
uses a single coefficient per sample. Channel-wise decomposition is applied
only to the PCA remainder: Rectified Flow uses a single joint pixel-space
basis, whereas SiT-XL/2 and SD3.5 Medium use separate PCA bases for each latent
channel. Thus, the reported rank \(K\) is the joint PCA rank for Rectified Flow
and the rank \emph{per latent channel} for SiT-XL/2 and SD3.5 Medium. The
explicit-direction coefficients and PCA coefficients are then concatenated to
form the training target for the coefficient head.

\section{Timestep-Schedule Optimisation Details}
\label{app:schedule_analysis}

This section expands the optional schedule-selection procedure from
Section~\ref{sec:time_schedule_optimisation}. It is applied to the Rectified
Flow and SiT configurations reported with optimised schedules; SD3.5 Medium
retains its built-in shifted schedule.

\subsection{Schedule Selection and Representative 8-Step Schedules}

Schedule optimisation is performed separately for each model and sampling-step
budget. The selected grid is therefore specific to the deployed configuration.
SD3.5 Medium retains its built-in shifted schedule, so this section reports the
selected 8-step grids for Rectified Flow CIFAR-10 and SiT-XL/2.

\begin{figure}[!htbp]
    \centering
    \includegraphics[width=\linewidth]{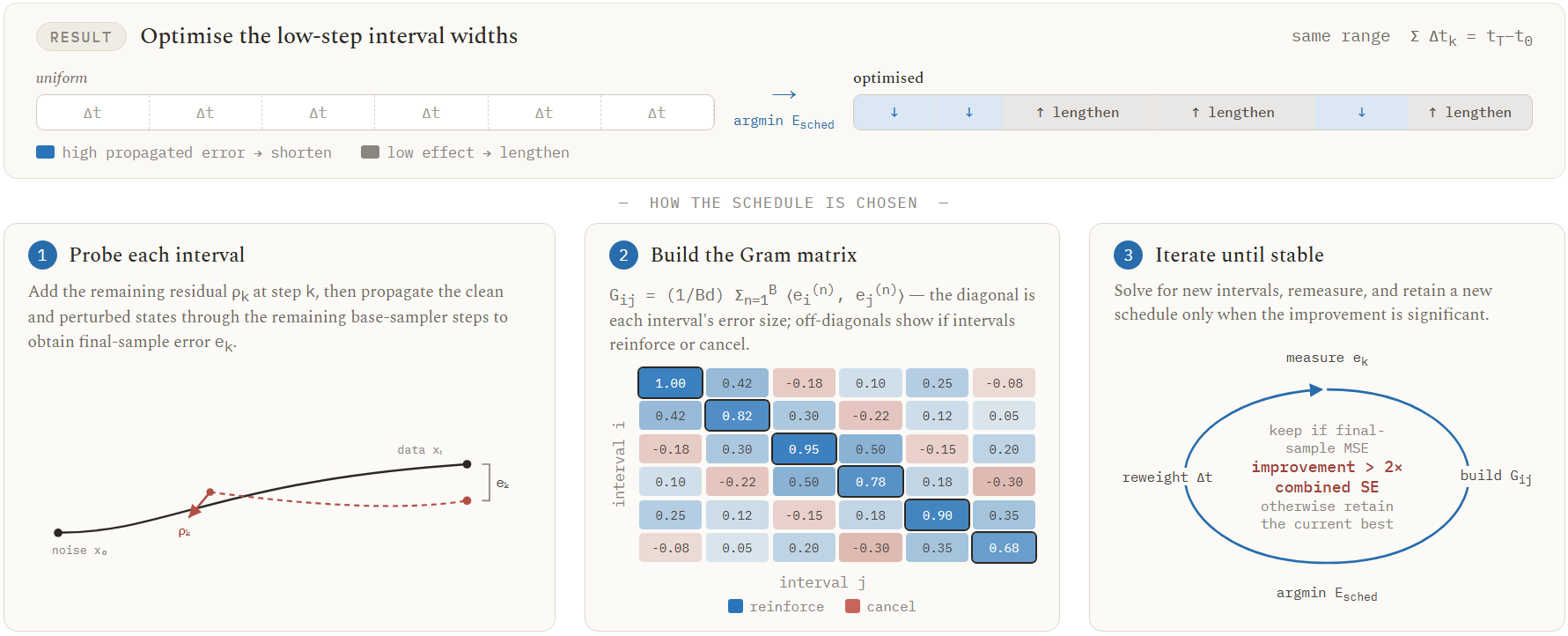}
    \caption{Overview of the optional timestep-schedule optimisation procedure.}
    \label{fig:time_schedule_optimisation}
\end{figure}

Figure~\ref{fig:time_schedule_optimisation} summarises the schedule-selection
procedure used to obtain these grids.

Both grids span
\[
    t_0=0.001
    \qquad\text{to}\qquad
    t_8=0.999.
\]
A uniform 8-step grid over this range would use
\[
    \Delta t_k^{\mathrm{uniform}}
    =
    \frac{0.998}{8}
    =
    0.124750.
\]

Table~\ref{tab:representative_8step_schedules} reports both the selected
sampling times \(t_k\) and the corresponding step sizes
\(\Delta t_k=t_{k+1}-t_k\). Values are rounded to six decimal places.

\begin{table}[!htbp]
    \centering
    \scriptsize
    \setlength{\tabcolsep}{5.5pt}
    \renewcommand{\arraystretch}{1.00}
    \caption{Selected sampling times and step sizes for the representative
        8-step schedules.}
    \label{tab:representative_8step_schedules}
    \begin{tabular}{ccccc}
        \toprule
        & \multicolumn{2}{c}{Rectified Flow CIFAR-10}
        & \multicolumn{2}{c}{SiT-XL/2} \\
        \cmidrule(lr){2-3}
        \cmidrule(lr){4-5}
        \(k\)
        & \(t_k\)
        & \(\Delta t_k\)
        & \(t_k\)
        & \(\Delta t_k\) \\
        \midrule
        0 & 0.001000 & 0.092182 & 0.001000 & 0.103061 \\
        1 & 0.093182 & 0.104293 & 0.104061 & 0.104207 \\
        2 & 0.197475 & 0.115650 & 0.208269 & 0.092542 \\
        3 & 0.313125 & 0.126756 & 0.300811 & 0.095130 \\
        4 & 0.439881 & 0.158832 & 0.395941 & 0.105806 \\
        5 & 0.598713 & 0.174227 & 0.501747 & 0.127641 \\
        6 & 0.772940 & 0.176058 & 0.629388 & 0.218872 \\
        7 & 0.948998 & 0.050000 & 0.848260 & 0.150740 \\
        8 & 0.999000 & --       & 0.999000 & --       \\
        \bottomrule
    \end{tabular}
\end{table}

Both schedules place timesteps more densely in the earlier part of the
trajectory than in the later part, suggesting that errors introduced early in
sampling have the greatest propagated effect on the final sample. For
Rectified Flow CIFAR-10, the step sizes generally increase from the beginning
to the penultimate update, allowing the later middle portion of the trajectory
to be traversed more coarsely. The final step is then reduced sharply to
\(0.050000\), suggesting that additional resolution is beneficial when the
corrector is omitted.

The SiT-XL/2 schedule shows a related but less regular pattern. Its first six
updates use step sizes close to or below the uniform width, placing six
intermediate timesteps before \(t=0.63\). The remaining part of the trajectory
is covered by two substantially larger updates. Thus, both models concentrate
sampling resolution earlier in the trajectory, while the short final step
appears only for Rectified Flow, indicating that the detailed allocation
remains model-dependent.

\subsection{Search Implementation}

The schedule search follows the procedure described in
Section~\ref{sec:time_schedule_optimisation}. Before the final coefficient
heads are trained, a lightweight probe computes oracle decomposition
coefficients directly from the known residual targets. These coefficients are
used only to measure the residual left by the compact representation; the probe
trajectory itself is advanced with the base sampler.

For probe trajectory \(n\), let \(\rho_k^{(n)}\) denote the unrepresented
residual left at step \(k\), using both correction stages except on the final
step, where only the predictor is deployed. Its propagated effect is measured
by adding it to the clean base endpoint and rolling both states through the
remaining base-sampler steps:
\[
    e_k^{(n)}
    =
    \mathcal{B}_{k+1:T}
    \!\left(x_{k+1}^{\mathrm{base}}+\rho_k^{(n)}\right)
    -
    \mathcal{B}_{k+1:T}
    \!\left(x_{k+1}^{\mathrm{base}}\right),
\]
where \(\mathcal{B}_{k+1:T}\) denotes the remaining base-sampler rollout.
The Gram matrix in Eq.~\eqref{eq:schedule_objective} is measured from these
propagated perturbations,
\[
    G_{ij}
    =
    \frac{1}{Bd}
    \sum_{n=1}^{B}
    \left\langle e_i^{(n)},e_j^{(n)}\right\rangle,
\]
with \(B\) probe trajectories and \(d\) scalar values in the sampling state.
This makes the diagonal terms individual propagated-error contributions and
the off-diagonal terms their interactions.

For a \(T\)-step schedule, the step sizes are parameterised by unconstrained
variables \(z_0,\ldots,z_{T-1}\). Given a minimum step size
\(\Delta t_{\min}\),
\[
    \Delta t_k
    =
    \Delta t_{\min}
    +
    \left(t_T-t_0-T\Delta t_{\min}\right)
    \frac{\exp(z_k)}
         {\sum_{j=0}^{T-1}\exp(z_j)}.
\]
This guarantees \(\Delta t_k\geq\Delta t_{\min}\) while preserving the fixed
total sampling-time range. The variables \(z_k\) are optimised using
L-BFGS-B.

Each proposed schedule is then evaluated directly on the probe trajectories.
For probe trajectory \(n\), the mean-squared combined propagated error is
\[
    m^{(n)}
    =
    \frac{1}{d}
    \left\|
        \sum_{k=0}^{T-1} e_k^{(n)}
    \right\|_2^2,
\]
where \(d\) is the number of scalar elements in the sampling state. The measured
propagated-error score is the mean over the \(B\) probe trajectories,
\[
    \widehat E
    =
    \frac{1}{B}
    \sum_{n=1}^{B} m^{(n)}.
\]
Lower values indicate a smaller propagated effect of the residual errors on
the final state.

The standard error of the measured propagated-error score is
\[
    \operatorname{SE}
    =
    \sqrt{
        \frac{1}{B(B-1)}
        \sum_{n=1}^{B}
        \left(
            m^{(n)}-\widehat E
        \right)^2
    }.
\]
A candidate replaces the current best schedule when
\[
    \widehat E_{\mathrm{best}}
    -
    \widehat E_{\mathrm{candidate}}
    >
    2
    \sqrt{
        \operatorname{SE}_{\mathrm{best}}^2
        +
        \operatorname{SE}_{\mathrm{candidate}}^2
    }.
\]
All schedule comparisons use matched probe trajectories with the same
initial-noise and conditioning seeds.

\begin{table}[H]
    \centering
    \scriptsize
    \setlength{\tabcolsep}{7pt}
    \renewcommand{\arraystretch}{1.00}
    \caption{Timestep-schedule search settings. All settings are shared across
        the optimised configurations except for the number of probe
        trajectories.}
    \label{tab:schedule_search_settings}
    \begin{tabular}{@{}p{0.39\linewidth}p{0.47\linewidth}@{}}
        \toprule
        Setting & Value \\
        \midrule
        Probe trajectories
        & \(4096\) for Rectified Flow CIFAR-10;
          \(2048\) for SiT-XL/2 \\

        Sampling-time range
        & \([10^{-3},1-10^{-3}]\) \\

        Minimum step size
        & \(0.001\) \\

        Error-scaling exponent
        & \(p=2\) \\

        Optimiser
        & L-BFGS-B \\

        Maximum optimisation passes
        & \(10\) \\

        Convergence criterion
        & \(\displaystyle
          \max_k
          \left|
          \log
          \left(
          \frac{\Delta t_k^{\mathrm{new}}}
               {\Delta t_k^{\mathrm{old}}}
          \right)
          \right|
          <0.02\) \\

        Acceptance criterion
        & Mean propagated-error improvement greater than twice the combined
          standard error \\
        \bottomrule
    \end{tabular}
\end{table}
\FloatBarrier

\section{Additional Experimental Results and Cost Details}
\label{app:additional_results}

This section collects supporting material that is useful for reproducibility or
design inspection without duplicating the main result tables: design-space
sweeps, an ImageNet-256 qualitative comparison, the prompts used for the SD3.5
qualitative comparison, and one-time preparation and coefficient-head costs.

\subsection{Design-space Ablations}
\label{app:rf_design_ablations}

\begin{figure}[H]
    \centering
    \includegraphics[width=\linewidth]{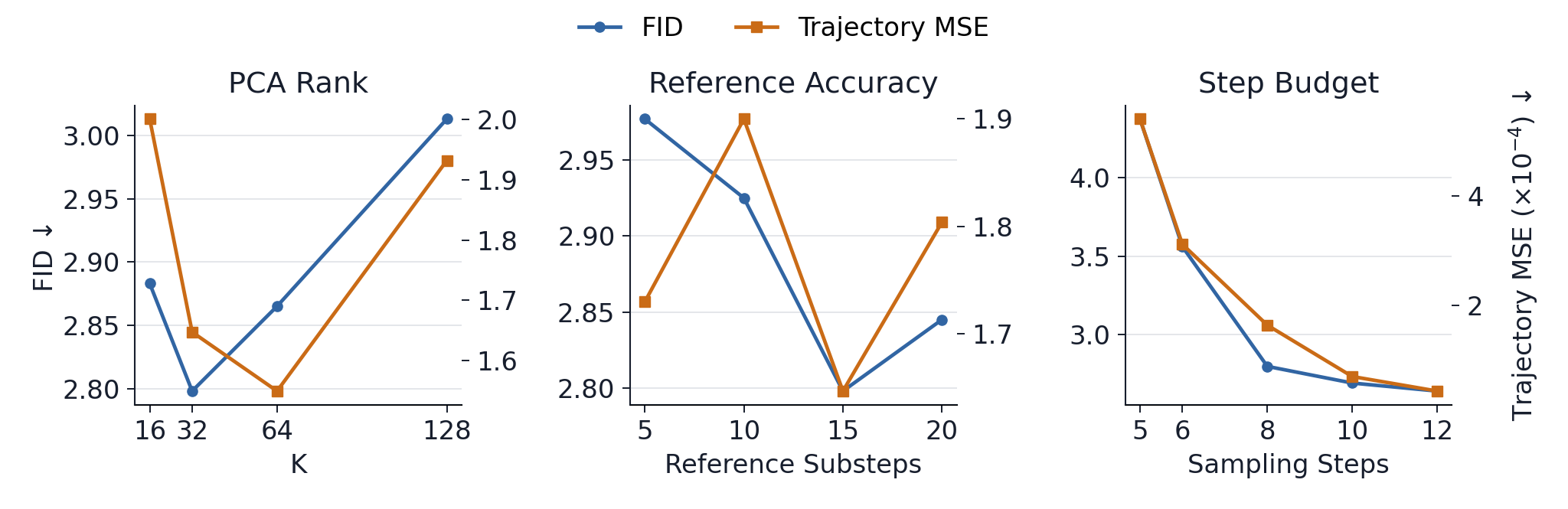}
    \caption{Design-space ablations.}
    \label{fig:rf_sweeps}
\end{figure}

We vary the PCA rank \(K\), the accuracy of the local
reference update, and the deployed sampling budget. The PCA-rank and
reference-accuracy sweeps use the 8-step TraceSampler configuration.
Figure~\ref{fig:rf_sweeps} summarises all three sweeps, while
Table~\ref{tab:rf_ablation_key_results} reports the PCA-rank and
reference-accuracy settings numerically. The deployed configuration is selected
primarily by FID, with IS and trajectory MSE used as complementary diagnostics.

\begin{table}[H]
    \centering
    \scriptsize
    \setlength{\tabcolsep}{7pt}
    \renewcommand{\arraystretch}{1.00}
    \caption{Key Rectified Flow CIFAR-10 design-space ablation results.}
    \label{tab:rf_ablation_key_results}
    \begin{tabular}{llrrr}
        \toprule
        Ablation
         & Setting
         & FID \(\downarrow\)
         & IS \(\uparrow\)
         & Trajectory MSE \(\downarrow\) \\
        \midrule
        \multirow{4}{*}{Reference accuracy}
         & 5 Heun substeps
         & 2.977
         & 8.526
         & \(1.730\times10^{-4}\) \\

         & 10 Heun substeps
         & 2.925
         & 8.508
         & \(1.900\times10^{-4}\) \\

         & 15 Heun substeps
         & \textbf{2.798}
         & 8.552
         & \(1.647\times10^{-4}\) \\

         & 20 Heun substeps
         & 2.845
         & 8.581
         & \(1.804\times10^{-4}\) \\

        \midrule
        \multirow{4}{*}{PCA rank}
         & \(K=16\)
         & 2.883
         & 8.530
         & \(2.001\times10^{-4}\) \\

         & \(K=32\)
         & \textbf{2.798}
         & 8.552
         & \(1.647\times10^{-4}\) \\

         & \(K=64\)
         & 2.865
         & 8.493
         & \(1.549\times10^{-4}\) \\

         & \(K=128\)
         & 3.013
         & 8.498
         & \(1.932\times10^{-4}\) \\
        \bottomrule
    \end{tabular}
\end{table}

Moderate settings are sufficient for both residual capacity and reference
accuracy. Increasing the PCA rank from \(K=16\) to \(K=32\) improves FID,
whereas larger ranks provide no further sample-quality gain. Although \(K=64\)
gives the lowest trajectory MSE, its FID and IS are slightly weaker than those
obtained with \(K=32\), showing that closer trajectory matching does not
translate monotonically into improved final-sample metrics.

A similar saturation is observed for reference accuracy. Fifteen Heun substeps
give the strongest FID result, while increasing the local reference to 20
substeps provides no further improvement. Performance therefore saturates once
the reference is sufficiently accurate to expose the coarse-step error.

\subsection{ImageNet-256 Qualitative Comparison}
\label{app:sit_qualitative}

\begin{figure}[H]
    \centering
    \includegraphics[width=0.77\linewidth]{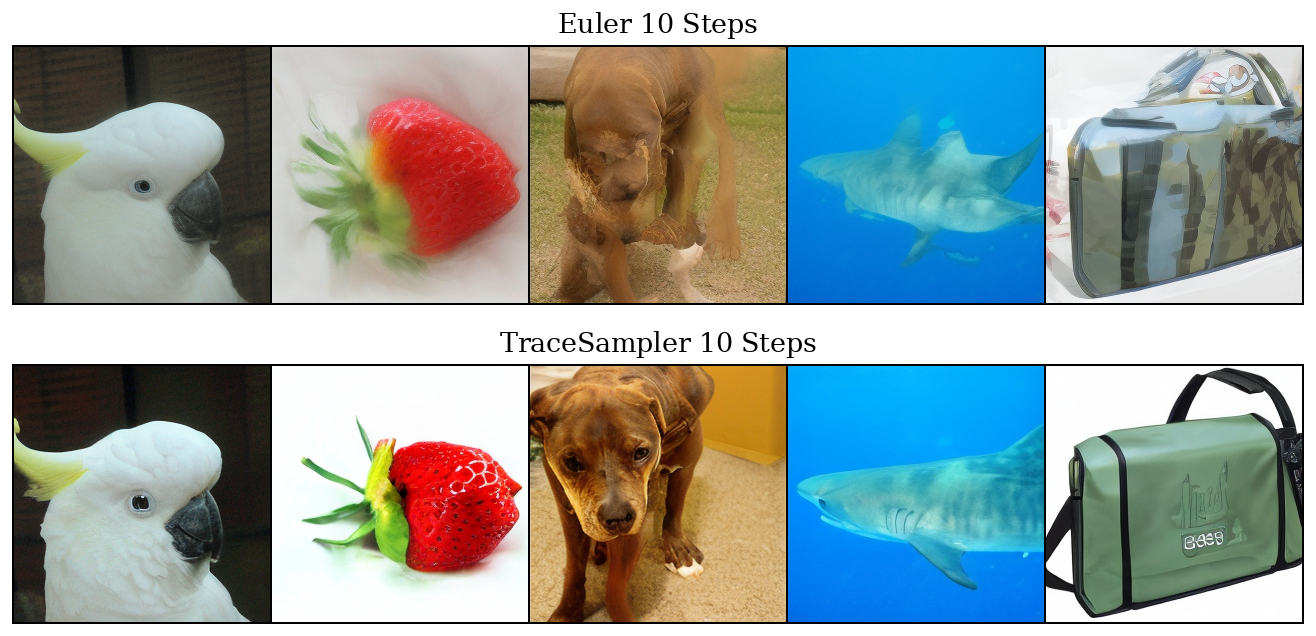}
    \caption{Qualitative comparison between Euler and TraceSampler on SiT-XL/2 ImageNet-256.}
    \label{fig:sit_comparison}
\end{figure}

Figure~\ref{fig:sit_comparison} compares Euler and TraceSampler at the same
10-step budget on SiT-XL/2 ImageNet-256.

\subsection{SD3.5 Qualitative Prompts}
\label{app:sd35_qualitative_prompts}

\begin{table}[!htbp]
    \centering
    \scriptsize
    \setlength{\tabcolsep}{6pt}
    \renewcommand{\arraystretch}{1.00}
    \caption{Prompts used in the SD3.5 Medium qualitative comparison.}
    \label{tab:sd35_qualitative_prompts}
    \begin{tabular}{@{}lp{0.82\linewidth}@{}}
        \toprule
        Label & Prompt \\
        \midrule
        P1 & A fluffy white cat sitting on a mossy mountain ledge under a huge full moon, soft starlight, crisp detailed fur, cinematic night landscape. \\
        P2 & A close-up portrait of a cyberpunk geisha on a rainy Tokyo rooftop, detailed expressive face, neon skyline, holographic lanterns, wet silk kimono, cinematic lighting. \\
        P3 & A masked knight in ornate silver armour standing inside a ruined cathedral, sword point resting on the stone floor in front of him, both hands on the hilt, glowing stained-glass light, misty air, cinematic atmosphere. \\
        P4 & A close-up portrait of an old lighthouse keeper in a storm, detailed weathered face, thick wool coat, lantern glow, rain on glass, cinematic lighting. \\
        \bottomrule
    \end{tabular}
\end{table}

\subsection{Preparation Measurements}
\label{app:preparation_measurements}

The CIFAR-10 and SiT-XL/2 experiments use an NVIDIA A100 80 GB GPU; SD3.5 Medium
uses an NVIDIA RTX PRO 6000 Blackwell 95 GB GPU. The table below reports the
one-time offline preparation measurements; inference measurements are reported
directly in the main paper.

\begin{table}[!htbp]
    \centering
    \scriptsize
    \setlength{\tabcolsep}{7pt}
    \renewcommand{\arraystretch}{1.00}
    \caption{Measured offline preparation time for 8-step TraceSampler configurations.}
    \label{tab:offline_runtime}
    \begin{tabular}{lrrr}
        \toprule
        Model & Schedule optimisation & Artefact fitting & Total \\
        \midrule
        RF CIFAR-10 & 0:26:57 & 2:32:42 & 2:59:39 \\
        SiT-XL/2 & 1:07:04 & 9:02:39 & 10:09:42 \\
        SD3.5 Medium & -- & 8:35:00 & 8:35:00 \\
        \bottomrule
    \end{tabular}
\end{table}

\subsection{Coefficient-Head Size}

The coefficient heads are lightweight in compute relative to the frozen
backbones but introduce additional persistent inference memory. Table~\ref{tab:head_parameters}
reports the parameter count of one predictor and one corrector head for each
model; all coefficient-head weights are stored in 16-bit precision at inference.

\begin{table}[!htbp]
    \centering
    \scriptsize
    \setlength{\tabcolsep}{8pt}
    \renewcommand{\arraystretch}{1.00}
    \caption{Parameters per TraceSampler coefficient head.}
    \label{tab:head_parameters}
    \begin{tabular}{lrr}
        \toprule
        Model & Predictor (M) & Corrector (M) \\
        \midrule
        Rectified Flow & 20.570 & 20.619 \\
        SiT-XL/2 & 25.30 & 25.35 \\
        SD3.5 Medium & 88.646 & 88.744 \\
        \bottomrule
    \end{tabular}
\end{table}

These counts describe separate step-specific heads rather than parameters added
to the frozen backbone. At each correction stage, only the corresponding
shallow MLP is evaluated on compact features. The additional parameters
therefore primarily contribute persistent weight storage and do not have the
same computational or activation-memory impact as an equivalent increase in
backbone size. The measured end-to-end inference-memory overhead is reported in
Table~\ref{tab:inference_cost}.

\section{Baseline and Evaluation Configurations}
\label{app:baseline_configurations}

This section records the sampler settings used for the matched-budget baselines
and the scoring protocols used throughout Section~\ref{sec:experiments}.

\subsection{Backbones and Conditioning}

The Rectified Flow experiments use the official CIFAR-10 1-Rectified-Flow
\emph{checkpoint\_8.pth}; ImageNet uses \emph{SiT-XL-2-256x256.pt}; and SD3.5
uses \emph{stabilityai/stable-diffusion-3.5-medium}. SiT uses CFG scale 1.5;
SD3.5 uses CFG scale 4.5 and no Skip-Layer Guidance. Rectified Flow and
SiT-XL/2 use 120-step Heun references; at 10 deployed steps this corresponds
to 12 reference substeps per coarse interval, while the 8-step RF ablation
uses 15. SD3.5 uses 8 Euler reference substeps at 8 steps and 6 at 10 steps,
yielding 64- and 60-interval references. For SD3.5 offline fitting,
conditioning prompts are drawn from a 30,000-caption bank built from COCO 2014
validation annotations using one caption per image for the first 30,000 image
IDs \citep{lin2014microsoft}. These captions are used only for conditioning:
residual targets are obtained from frozen-backbone rollouts, with no target
images or image-derived supervision used during TraceSampler fitting. GenEval
and HPSv2.1 use their respective official prompt sets.

\subsection{Sampler Baselines}

\begin{table}[!htbp]
    \centering
    \scriptsize
    \setlength{\tabcolsep}{6pt}
    \renewcommand{\arraystretch}{1.12}
    \caption{Configurations used for the sampler baselines.}
    \label{tab:baseline_details}
    \begin{tabular}{@{}p{0.20\linewidth}>{\raggedright\arraybackslash}p{0.73\linewidth}@{}}
        \toprule
        Baseline & Configuration \\
        \midrule
        IIA--Euler
        & Basic \(r=1\) IIA adapted to the flow-matching update. Two global
          coefficients combine the current and previous Euler increments.
          Calibration uses 512 trajectories, three fine-Euler substeps per
          coarse step, and ridge regularisation of \(10^{-10}\). \\

        UniPC
        & Multistep predictor with order 3, the \texttt{bh2} variant, log-SNR
          spacing, \texttt{data\_prediction}, lower-order final steps, and one
          model evaluation per deployed step. \\
        \bottomrule
    \end{tabular}
\end{table}

\subsection{Evaluation Protocols}

\begin{table}[H]
    \centering
    \scriptsize
    \setlength{\tabcolsep}{5pt}
    \renewcommand{\arraystretch}{1.13}
    \caption{Evaluation sets and scoring configurations.}
    \label{tab:evaluation_protocols}
    \begin{tabular}{@{}p{0.20\linewidth}>{\raggedright\arraybackslash}p{0.73\linewidth}@{}}
        \toprule
        Evaluation & Configuration \\
        \midrule
        CIFAR-10
        & \(50000\) generated samples. FID uses 2048-dimensional Inception-V3
          features against CIFAR-10 training-set statistics; IS uses ten
          splits. \\

        ImageNet-256
        & \(50000\) class-conditional samples. FID and IS use the
          \texttt{VIRTUAL\_imagenet256\_labeled.npz} reference statistics and
          ten IS splits. \\

        COCO captions
        & One image for each of \(30000\) captions. The evaluation reports
          ImageReward-v1.0, Inception FID and IS, raw CLIP-L/14 image--text
          cosine similarity, reported as CLIP similarity, and trajectory MSE
          against the matched higher-step reference. \\

        GenEval
        & Standard 553-prompt metadata with four images per prompt, CFG \(4.5\),
          and the official MMDetection-2 scoring pipeline. \\

        HPSv2.1
        & Official anime, concept-art, paintings, and photo prompt subsets, with
          one image per retained prompt and the mean of the four subset scores.
          Prompts exceeding 77 tokens for any SD3 text encoder are excluded. \\
        \bottomrule
    \end{tabular}
\end{table}

\end{document}